\PassOptionsToPackage{backref=page}{hyperref}
\documentclass{article}

\usepackage{microtype}
\usepackage{graphicx}
\usepackage{booktabs} \usepackage{tabularx} 

\usepackage{subcaption}

\usepackage{hyperref}

\renewcommand*{\backrefalt}[4]{\ifcase #1 No citations.\or
Cited on page #2.\else
Cited on pages #2.\fi
}

\usepackage[accepted]{preprint}

\usepackage{amsmath}
\usepackage{amssymb}
\usepackage{mathtools}
\usepackage{amsthm}
\usepackage[thinc]{esdiff}

\usepackage[capitalize,noabbrev]{cleveref}

\theoremstyle{plain}

\theoremstyle{definition}

\theoremstyle{remark}

\usepackage{multirow}
\usepackage{rotating}

\newcommand{\acr}[1]{{{\textsc{#1}}}}
\newcommand{\ourmodel}[0]{GraphToken}
\newcommand{\method}[1]{\acr{#1}}
\newcommand{\eg}[0]{\emph{e.g.},}
\newcommand{\ie}[0]{\emph{i.e.},}
\newcommand{\task}[1]{{{\texttt{#1}}}}

\usepackage[textsize=tiny]{todonotes}

\usepackage{enumitem}
\setlist[itemize]{noitemsep}
\setlist{nosep}

\usepackage{tikz}
\usepackage{pgfplots}
\pgfplotsset{compat=1.16}
\usepgfplotslibrary{fillbetween, groupplots, colorbrewer, colormaps}

\definecolor{cycle1}{RGB}{235,172,35}
\definecolor{cycle2}{RGB}{184,0,88}
\definecolor{cycle3}{RGB}{0,140,249}
\definecolor{cycle4}{RGB}{0,110,0}
\definecolor{cycle5}{RGB}{0,187,173}
\definecolor{cycle6}{RGB}{209,99,230}
\definecolor{cycle7}{RGB}{178,69,2}
\definecolor{cycle8}{RGB}{255,146,135}
\definecolor{cycle9}{RGB}{89,84,214}
\definecolor{cycle10}{RGB}{0,198,248}
\definecolor{cycle11}{RGB}{135,133,0}
\definecolor{cycle12}{RGB}{0,167,108}
\definecolor{cyclegray}{RGB}{189,189,189}

\begin{document}

\twocolumn[
\icmltitle{Let Your Graph Do the Talking: Encoding Structured Data  for LLMs}

\begin{icmlauthorlist}
\icmlauthor{Bryan Perozzi}{goog}
\icmlauthor{Bahare Fatemi}{goog}
\icmlauthor{Dustin Zelle}{goog}
\icmlauthor{Anton Tsitsulin}{goog}
\\
\icmlauthor{Mehran Kazemi}{goog}
\icmlauthor{Rami Al-Rfou}{waymo}
\icmlauthor{Jonathan Halcrow}{goog}
\end{icmlauthorlist}

\icmlaffiliation{goog}{Google Research}
\icmlaffiliation{waymo}{Waymo Research}

\icmlcorrespondingauthor{Bryan Perozzi}{bperozzi@acm.org}

\vskip 0.3in
]

\printAffiliationsAndNotice{}

\begin{abstract}

How can we best encode structured data into sequential form for use in large language models~(LLMs)?
In this work, we introduce a parameter-efficient method to explicitly represent structured data for LLMs.
    Our method, \ourmodel{}, learns an encoding function to extend prompts with explicit structured information.
    Unlike other work which focuses on limited domains (\eg{} knowledge graph representation), our work is the first effort focused on the general encoding of structured data to be used for various reasoning tasks.
    We show that explicitly representing the graph structure allows significant improvements to graph reasoning tasks.
    Specifically, we see across the board improvements - up to 73\% points -  on node, edge and, graph-level tasks from the GraphQA benchmark.

\end{abstract}

\section{Introduction}
\label{introduction}

There has been an explosion of recent excitement around using LLMs~\cite{vaswani2017attention,devlin2018bert,radford2018improving,raffel2020exploring,fewshot,touvron2023llama,zhao2023survey} to represent, process, and analyze textual data.
These models typically take sequential text as their input but recent work has extended inputs to spatial and temporal modalities (\eg{} image~\cite{chen2022pali} and video~\cite{arnab2021vivit}).

\begin{figure}[t]
    \centering
    \includegraphics[width=0.8\linewidth]{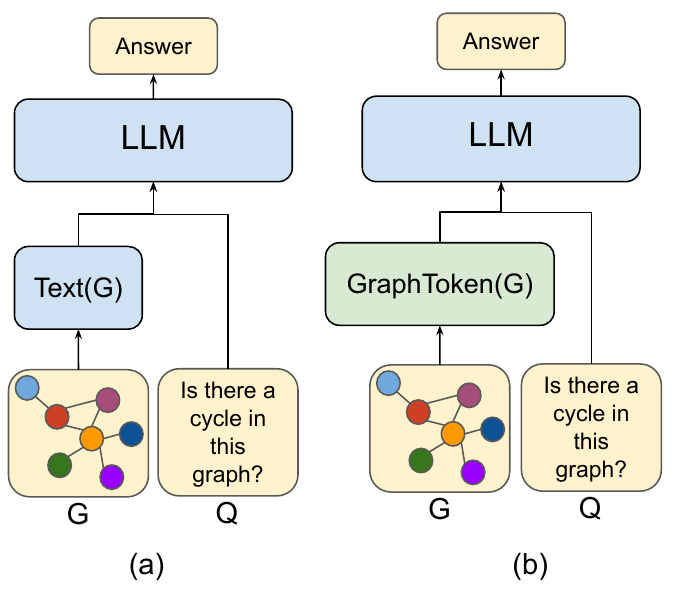}
    \caption{Graph encoding options for a frozen LLM.
    \textbf{a)} Fixed encoding, \eg{} \cite{fatemi2023talk,wang2023can,stechly2023gpt}, 
    \textbf{b)} This work proposes using \ourmodel{}, a learned graph prompt function to explicitly encode graphs in a parameter efficient way.}
    \label{fig:encoding_options}
\end{figure}

Despite their success, current realizations of LLMs have noticeable problems -- including a tendency to generate outputs which are untrue or unsupported by their prompt, commonly referred to as \textit{hallucinations} \cite{wang2023survey}.
Another intimately related issue is the problem of \textit{freshness}, where the knowledge required to answer a query exists only after an LLM's training date~\cite{vu2023freshllms}.
One mitigation for these problems is through the enrichment of the prompt with additional factual and fresh data.
As \citet{kadavath2022language} showed, when LLMs are supplied with new and supporting information, they are capable of adapting their parametric beliefs to effectively incorporate new evidence. 

An automatic way to enrich the input context of an LLM with factual and fresh information is through Retrieval Augmented Generation~(RAG)~\cite{khandelwal2019generalization,guu2020retrieval}.
RAG works by augmenting the prompt with additional relevant, factual and fresh information. Sources for RAG might include web searches or private databases. 
Often this information is in the form of \textit{structured data} -- data that has complex dependencies between different, \textit{discrete} parts of the whole. 
For example, private relational databases, social networks, or molecules all have relational information between their discrete data items. 

Structured data is ubiquitous in the real world -- it surrounds our daily lives -- and understanding how to represent this data optimally for its inclusion in LLMs is a crucial research question.
The predominant mode of encoding structured data for LLMs is to use various types of \textit{hand-crafted}, text-based serialization~\cite{guo2023gpt4graph,wang2023can,stechly2023gpt} (See \Cref{fig:encoding_options} (a)).
This approach can impose significant decoding complexity for the language model: from any text description, the model must first correctly decode and understand the structure before it can utilize the information.
Recently, \citet{fatemi2023talk} demonstrated that pure text representations of structured data are insufficient for graph reasoning with LLMs.
They show that LLMs are not able to utilize structure efficiently when posed with common reasoning tasks that are easily answered by classical graph algorithms.
This highlights the need to explore better and more expressive ways of representing structured data to an LLM.

In this paper, we propose \ourmodel{} (\Cref{fig:encoding_options} (b)), a parameter-efficient method for representing structured data for LLMs.
Pre-training LLMs on text corpora closely related to the desired reasoning task can enhance performance, but it can be computationally expensive, particularly for larger models. Additionally, fine-tuning requires domain-specific data and human expertise, further increasing costs. 
Inspired by recent advancements in parameter-efficient fine-tuning~\cite{lester2021power,xu2023parameter}, our method, \ourmodel{}, learns an encoding function that generates fine-tuned soft-token prompts. The soft-token prompt extends a textual prompt with explicit \ourmodel{} encoded structural information, allowing us to train only a trivial number of \ourmodel{} parameters when compared to the total LLM parameter budget.  

Our work is the first to develop parameter-efficient encoders specifically for general reasoning tasks on structured data.
We demonstrate that explicitly representing structure leads to significant improvement on the comprehensive GraphQA benchmark \cite{fatemi2023talk}.

\textbf{Our Contributions.}
We propose the following innovations:
\begin{itemize}
    \item \ourmodel{}, a novel parameter-efficient encoder for structured data inclusion in LLMs.
    \item Extensive experiments on various graph reasoning tasks showing that our method significantly improves LLM capabilities.
    \item Analysis demonstrating that the \ourmodel{} encoder generalizes to both unseen tasks and graphs.
\end{itemize}

\section{Background}

We introduce the related work in LLMs, prompting methods, Graph Neural Networks (GNNs), graph encoders, and graph models combined with LLMs.

\subsection{Large Language Models}

\paragraph{Pre-Trained Large Language Models (LLMs):} Language models \cite{rosenfeld2000two,zhao2023survey} are probabilistic models that assign probabilities to sequences of words by breaking the probability of a sequence into the product of the probabilities of the next tokens given the previous ones.
While earlier models were mainly based on N-gram models \cite{jurafsky2021}, newer models adopted neural approaches with the advent of distributed word representations \cite{bengio2000neural,mikolov2013efficient}.
The increased power offered by the neural language models and the increase in model and dataset sizes has led to a new learning paradigm where large language models (LLMs) are pre-trained in an unsupervised way on massive amounts of textual data and are used as base (foundation) models \cite{devlin2018bert,radford2019language}. For each downstream application, the base model is fine-tuned on small amounts of task-specific data to adapt the model to the task.

\paragraph{Parameter-Efficient Fine-Tuning:} With the rapid growth in the number of parameters for the state-of-the-art LLMs \cite{achiam2023gpt,team2023gemini} fine-tuning for each downstream task has become prohibitively expensive in both time and resources.
The goal of parameter-efficient fine-tuning (PEFT) \cite{xu2023parameter} is to adapt models to new tasks by updating only a small number of (possibly new) parameters.
There are a few dominant PEFT approaches:
\begin{itemize}
    \item \textit{Adapter-based} approaches \cite{houlsby2019parameter,he2021effectiveness} hold the LLM parameters frozen and add new trainable parameters to various parts of the model, with the main differentiating factor between different approaches being where the adapter parameters are added.
    \item \textit{LoRA} and its variants \cite{hu2021lora,edalati2022krona,valipour2022dylora} similarly hold the LLM parameters frozen and add new trainable parameters, however these trainable parameters are added to the frozen LLM parameters such that the fine-tuned LLM is identical in architecture to the initial LLM, but with only those added parameters update.
    \item Partial fine-tuning and partial masking approaches \cite{zhao2020masking,zaken2021bitfit} only fine-tune or mask a subset of the LLM parameters -- no new parameters are introduced.
    \item Finally, \textit{soft-prompt} approaches \cite{li2021prefix,lester2021power} prepend tokens with learnable parameters to the beginning of the LLM input or to the beginning of every LLM layer -- like adapter-based and LoRA approaches, they hold the actual LLM parameters frozen.
\end{itemize}

Our work falls under the umbrella of soft-prompt approaches but can be extended to other PEFT approaches as well.
Most relevant to our work is the work of \citet{levine2022standing}, where the input is fed into a separate trainable neural network to produce the soft-prompt. We extend this to encoding structured data input via a GNN to produce the LLM soft-prompt. 

\subsection{Graph Encoding with Neural Networks}

The field of graph representation learning \cite{chami2022machine} seeks ways to represent structured data (\ie{} discrete and relation) in a continuous domain -- typically for use in a downstream machine learning task.
While seminal work like DeepWalk \cite{perozzi2024deepwalk} popularized the \emph{node embedding} problem, later work utilized GNNs to generalize and learn representations of the entire graph (\emph{graph embeddings}).
Many approaches learning graph representations (\emph{node} or \emph{graph} embeddings) have followed \cite{tsitsulin2018sgr,xie2022self}.

\subsection{Graphs and LLMs}

The confluence of graph representation learning and reasoning with LLMs is a rapidly growing field of research -- like language, structured data surrounds us but, unlike LLM input, it is not sequential.
Some of the first graphs in the literature are knowledge graphs as in ~\cite{agarwal2020knowledge}, where the retrieval corpus of a retrieval LLM is augmented with text-encoded knowledge graphs. 
\citet{ye2023natural} utilize instruction fine-tuned LLMs for node classification.
Similarly, \citet{chen2023exploring} leverage LLMs to enhance graph learning models by incorporating rich text attributes. \citet{wang2023can} showed that language models demonstrate preliminary abilities for graph reasoning tasks.
Later, \citet{fatemi2023talk} proposed GraphQA -- a comprehensive benchmark to systematically evaluate models for graph reasoning tasks -- finding that graph reasoning tasks are currently difficult and that scaling laws do not seem to apply. 
Finally, while there is a growing body of work in pre-training, fine-tuning, and prompt-tuning with GNNs by themselves ~\cite{fang2023universal,liu2023graphprompt}, the research, though conceptually similar, differs crucially from our work. GNN-based approaches lack the textual understanding capabilities that are central to the integration of LLMs with graph learning and reasoning.

\begin{figure*}[t]
    \centering
    \includegraphics[width=0.75\textwidth]{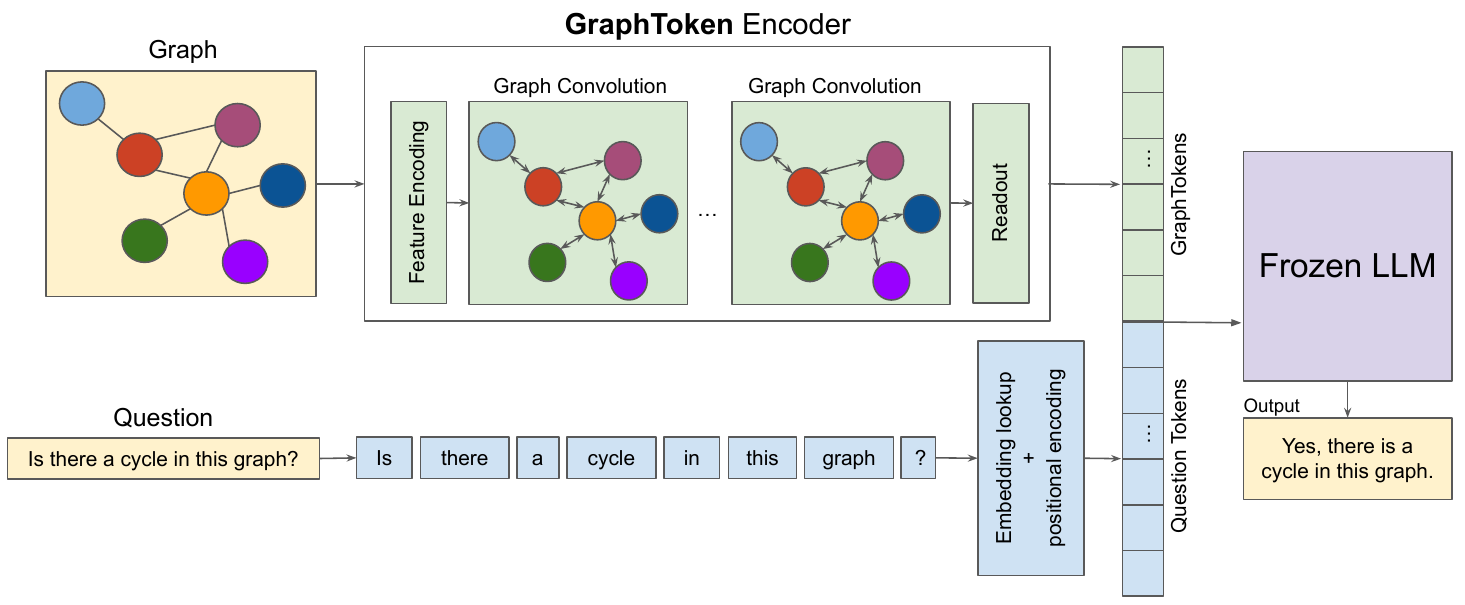}
\caption{A visual overview of the architecture of \ourmodel{}. The framework takes a graph and a corresponding question as input. The graph encoder takes the graph and generates graph tokens. The question is tokenized and embedded to question tokens. A frozen LLM leverages the graph and question tokens to generate an answer.}
    \label{fig:model_architecture}
\end{figure*}

\section{\ourmodel{}}
When considering how to pass structured data to an LLM there are largely two families of options: (1) encoding it as lexical tokens for LLM embedding as in \cite{fatemi2023talk} or (2) encoding it directly to a continuous representation via a neural network -- skipping any LLM token embedding.  
While representing a graph as a sequence of lexical tokens has benefits in terms of interpretability,
there is often no clear choice in what order to sequentially write the structured data.
We believe a text encoding of structured data prohibits rich, concise, and expressive representations.
Instead, our method eschews representing a graph in text in favor of directly producing -- using a GNN as an encoder -- the continuous representations for the LLM input. We refer to these new graph encoder learned soft-tokens in the LLM embedding space as ``graph tokens.''
 
To maintain the reasoning and language capabilities of the LLM, we freeze its parameters and teach the graph encoder to align its output representations with the LLM embedding space: we learn only those parameters of the graph encoder during the training process.
This reduces computational requirements significantly (graph encoder parameters constitute a trivial sum compared to the LLM).
During our tests, the LLM is prompted with the output of the graph encoder and a task about
the graph, for example: \textit{`Does this graph contain a cycle?'}.
As such, the quality of the results is purely a function of how well the graph encoder represents the answer to the task and how well the LLM interprets
that output.

\subsection{Architecture}

An overview of the architecture is provided in Figure \ref{fig:model_architecture}.
At a high level, our model only has two components.  First, the graph encoder takes a graph as input and outputs a fixed number of token embeddings.
These tokens are then prepended to the sequence of initial token embeddings in the prompt for an LLM, which is then decoded to produce an answer as text. 

\textbf{Graph Encoder}. GNN models range from simple averaging methods to complex models with multi-headed attention.
Thus there are a wide variety of graph representations possible.
We suspect that some of these representations are more suited to be consumed by an LLM.
Therefore, we conducted a thorough study that includes several well-known graph encoder choices in Section \ref{sec:encoder_design}.
Our graph encoder takes the relational structure of the graph as input, using some form of graph positional encoding as node features (either learned, Laplacian, or a combination thereof) - see Section \ref{subsec:gnn_features} for details.)
Next, we apply a GNN to obtain a representation of the graph, which we read out using one of a few different techniques techniques depending on the task.
\begin{itemize}
    \item For \textbf{graph-level} tasks (\eg{} \task{cycle check}) we do global pooling for readout, taking the \textit{mean} or \textit{sum} of the representations over all of the nodes. 
    \item For \textbf{node-level} tasks (\eg{} \task{node degree}) we separately output the representation of each node.  This can be optionally concatenated with a graph-level pooling.
    \item For \textbf{edge-level} tasks (\eg{} \task{edge existence}), we use a global representation or the two node-level representations concatenated.
\end{itemize}
We note that the exact option for readout used (e.g. mean or sum pooling) is a hyper-parameter chosen during model selection.
Whichever the readout technique, this representation is then projected onto
the space of tokens used by the LLM with a final dense layer. 

\textbf{LLM}. For the experiments in the paper we use PaLM 2\cite{anil2023palm}, however, our method generalizes to nearly any LLM in use today.
For our purposes, any language model which can accept a sequence of token embeddings and produce text is acceptable, so long as
it is possible to compute a gradient with respect to part of the input sequence.

\begin{table*}[t]
\caption{
Results comparing \ourmodel{} against prompt engineering and soft prompting on graph reasoning tasks using the GraphQA$_{\text{Test}}$ benchmark \cite{fatemi2023talk}, by simple accuracy.  We see that \ourmodel{} substantially improves LLM performance on all graph, node, and edge-level tasks. The first best result for each task is highlighted in bold and the second best result is highlighted with an underline.}\label{table:main_results}
\footnotesize
\resizebox{\textwidth}{!}{\setlength{\tabcolsep}{4pt}
\begin{tabular}{l|cccc|cc|ccc} 
\toprule
& \multicolumn{4}{c}{\textbf{Graph Tasks}} & \multicolumn{2}{c}{\textbf{Node Tasks}} & \multicolumn{3}{c}{\textbf{Edge Tasks}} \\
\cmidrule(lr){2-5} \cmidrule(lr){6-7} \cmidrule(lr){8-10}
\textbf{Method} & \textbf{Node count} & \textbf{Edge count} & \textbf{Cycle check} & \textbf{Triangle counting} & \textbf{Node degree} & \textbf{Connected nodes} & \textbf{Reachability} & \textbf{Edge existence} & \textbf{Shortest path}\\ \midrule
\acr{zero-shot} & 0.217 & 0.124 & 0.760 & 0.015 & 0.140 & 0.147 & \underline{0.849} & 0.445 & 0.115 \\
\acr{zero-cot} & 0.146 & 0.094 & 0.323 & 0.127 & 0.104 & 0.088 & 0.735 & 0.335 & 0.336 \\
\acr{few-shot} & 0.253 & 0.120 & 0.374 & 0.030 & 0.174 & 0.124 & 0.794 & 0.368 & 0.227 \\
\acr{cot} & 0.276 & \underline{0.128} & 0.580 & 0.081 & 0.292 & 0.131 & 0.452 & 0.428 & 0.386 \\
\acr{cot-bag} & \underline{0.269} & 0.125 & 0.521 & 0.081 & \underline{0.280} & \underline{0.158} & 0.452 & 0.373 & 0.404 \\
\acr{soft-prompt} & 0.056 & 0.018 & \underline{0.832} & \underline{0.162} & 0.098 & 0.068 & 0.838 & \underline{0.544} & \underline{0.462} \\
\midrule
\textbf{\ourmodel{}} & \textbf{0.996} & \textbf{0.426} & \textbf{0.956} & \textbf{0.348} & \textbf{0.962} & \textbf{0.264} & \textbf{0.932} & \textbf{0.738} & \textbf{0.638} \\
\bottomrule
\end{tabular}
}
\end{table*}

\subsection{Training procedure}

Our training procedure is very similar to that used by soft prompting methods \cite{lester2021power}.
The training input consists of triples $(G, T, A)$ where $G$ is a graph structure, $T$ is a statement or question describing the task  (\eg{} `Does this graph contain a cycle?' for \task{cycle check}) and $A$ is the ground truth answer (`Yes, there exists a cycle in this graph.'). 

In the forward pass, we compute the augmented query $Q = \mathcal{E}(G) || \mathcal{T}(T)$, concatenating the \ourmodel{} encoding of the graph $\mathcal{E}(G)$ with the initial embedding of the task textual representation, $\mathcal{T}(T)$. 

We train by optimizing the final LLM perplexity (total log-likelihood), $\mathcal{L}(A \mid Q)$, of the expected answer $A$ with respect to the augmented query, $Q$. 
We minimize this loss, back-propagating the gradient of $\mathcal{L}$ with respect to $\mathcal{E}(G)$ to the parameters of the \ourmodel{} encoder -- keeping all LLM parameters frozen.  We use the Lion optimizer \cite{chen2023symbolic} with a learning rate $\alpha = 0.05$.

\section{Experiments}
In this section, we summarize the key experiments conducted with \ourmodel{}.
We begin by highlighting some of the most exciting results from our analysis here:
\begin{itemize}
    \item \textbf{R1}: \ourmodel{} demonstrates superior performance compared to established baselines across a comprehensive range of graph reasoning tasks, including graph-level, node-level, and edge-level tasks.
    \item \textbf{R2}: The performance of different graph convolution architectures varies across tasks. This highlights the importance of carefully choosing the right architecture for the specific graph reasoning problem at hand.
    \item \textbf{R3}: By intentionally breaking equivariance, we enhance \ourmodel{}'s graph reasoning capabilities. 
\end{itemize}

\paragraph{Datasets.}
We conduct our experiments on the graph reasoning tasks proposed in GraphQA~\cite{fatemi2023talk}. This dataset presents multiple graph reasoning problems with different difficulty levels. These tasks can be categorized as follows.
\begin{itemize}
    \item \textbf{Graph-level.}
    \task{node count} (counting the number of nodes in a graph), \task{edge count} (counting the number of edges in a graph),
    \task{cycle check} (determining whether a graph contains a cycle), and
    \task{triangle counting} (counting the number of triangles in a graph).
    
    \item \textbf{Node-level.}
    \task{node degree} (calculating the degree of a given node in a graph), and
    \task{connected nodes} (finding all the nodes that are connected to a given node in a graph),
    
    \item \textbf{Edge-level.}
    \task{reachability} (finding if there is a path from one node to another),
    \task{edge existence} (whether a given edge exists in a graph, and
    \task{shortest path} (finding the length of the shortest path from one node to another).
\end{itemize}
      
\paragraph{Setting.}
We implemented \ourmodel{} in Tensorflow~\citep{tensorflow} using the TF-GNN library~\citep{ferludin2023tfgnn}.
The LLM used in our experiments is the instruction-fine-tuned Flan~\citep{flan} checkpoint of PaLM 2 S~\citep{anil2023palm}. Experiments were carried out on Google TPUv3 and TPUv5e \cite{tpu}.
Model selection was performed by evaluating performance on GraphQA$_{\text{Train}}$

\subsection{Experiment 1: \ourmodel{} Performance}
\label{sec:exp1_graphtoken}
In this experiment, we rigorously evaluate the performance of \ourmodel{} against the following comprehensive set of established baselines:

\begin{itemize}
    \item{\method{zero-shot}.} In this approach, the model is given a task description and immediately asked to produce the desired output. No additional examples or demonstrations are provided.
    
    \item{\method{few-shot}.} This approach provides the model with a few examples of the task and their desired outputs~\citep{fewshot}. Unlike traditional training, these examples are included directly in the prompt, allowing the model to learn and adapt during the inference.
    
    \item{\method{CoT}.} Chain-of-thought~(CoT) prompting~\citep{wei2022chain} provides examples each showing step-by-step reasoning, teaching the LLM to generate its own thought processes for tackling new tasks. 
    
    \item{\method{zero-cot}.} Zero-shot CoT~\citep{kojima2022large} builds upon Chain-of-Thought (CoT) prompting by eliminating the need for training examples.  The LLM generates its own step-by-step reasoning process using a simple trigger phrase like ``Let's think step by step''.
    
    \item{\method{cot-bag}.} BAG prompting~\citep{wang2023can} extends \method{cot} to improve the performance of LLMs on graph-related tasks by appending ``Let's construct a graph with the nodes and edges first'' to the prompt.
    
    \item{\method{soft-prompt}.} This approach uses the standard soft prompt tuning of \citet{lester2021power}. It optimizes a global \textit{static} prompt which is shared across problem instances to improve task performance. Unlike our proposed method, it does not have access to the graph information, making the results of this approach equivalent to that of a majority classifier.
\end{itemize}

\paragraph{Results.}
\begin{table*}[t]
\caption{
Study of individual graph encoder performance on GraphQA$_{\text{Test}}$ tasks. Note that there is `no free lunch' here -- no single encoder examined dominates across all tasks.}\label{table:encoder_choice}
\footnotesize
\resizebox{\textwidth}{!}{\setlength{\tabcolsep}{4pt}
\begin{tabular}{ll|cccc|cc|ccc} 
\toprule
 & & \multicolumn{4}{c}{\textbf{Graph Tasks}} & \multicolumn{2}{c}{\textbf{Node Tasks}} & \multicolumn{3}{c}{\textbf{Edge Tasks}} \\
\cmidrule(lr){3-6} \cmidrule(lr){7-8} \cmidrule(lr){9-11}
& \textbf{Method} & \textbf{Node count} & \textbf{Edge count} & \textbf{Cycle check} & \textbf{Triangle counting} & \textbf{Node degree} & \textbf{Connected nodes} & \textbf{Reachability} & \textbf{Edge existence} & \textbf{Shortest path} \\
\midrule
\scriptsize{\multirow{5}{*}{\begin{sideways}Non-linear\end{sideways}}} & GCN & 0.746 & 0.056 & \textbf{0.964} & 0.208 & 0.264 & \textbf{0.264} & 0.918 & 0.68 & 0.604 \\
& GIN & 0.704 & 0.052 & 0.898 & 0.194 & 0.252 & 0.18 & 0.902 & 0.65 & 0.586 \\
& MPNN & 0.792 & \underline{0.368} & 0.956 & \textbf{0.348} & \textbf{0.962} & \underline{0.25 }& 0.934 & 0.648 & \textbf{0.638} \\
& HGT & 0.252 & 0.084 & 0.934 & 0.234 & 0.266 & 0.184 & \textbf{0.944} & \underline{0.718} & 0.6 \\
& MHA & \underline{0.912} & 0.264 & \underline{0.962} & \underline{0.266} & \underline{0.552} & 0.244 & 0.932 & \textbf{0.738} & \underline{0.608} \\
\midrule
\scriptsize{\multirow{2}{*}{\begin{sideways}Linear\end{sideways}}} & Node Set & \textbf{0.996} & 0.080 & 0.948 & 0.198 & 0.19 & 0.118 & \underline{0.942} & 0.596 & 0.568 \\
& Edge Set & 0.618 & \textbf{0.426} & \textbf{0.964} & 0.228 & 0.22 & 0.096 & 0.904 & 0.592 & 0.568 \\
\bottomrule
\end{tabular}
}
\end{table*} The results of this experiment, summarized in \Cref{table:main_results}, demonstrate that \ourmodel{} significantly outperforms existing methods on all graph, node, and edge-level tasks. 
While \method{soft-prompt} achieves the second best score on some tasks, this is mainly due to its ability to predict majority labels. For example, $82\%$ of the questions in \task{cycle check} are about existent cycles. Similarly, $54\%$ of the questions are about non-existent edges in \task{edge existence}.

\subsection{Experiment 2: Encoder Design}
\label{sec:encoder_design}
From the results in Table \ref{table:main_results}, we can see that graph encoders can significantly improve a LLM's capability on graph reasoning tasks.  
However the choice of graph encoders has a significant effect on task performance.
Here we study how different architecture choices affect the quality of the graph representation for a language model's use, including the choices of the graph convolution, the features available to the network, and the hyper-parameters.

\subsubsection{Choice: Graph Convolution}
\label{sec:exp2_graph_conv}
This experiment investigates the impact of graph convolution choice on the performance of \ourmodel{}. We evaluate the following diverse set of encoders:

\begin{itemize}
    \item \textbf{Graph Convolutional Network (GCN)}: One of the earliest GNNs, this model does mean pooling of neighbor features, followed by a non-linear transformation. \cite{kipf2017semisupervised}
    \item \textbf{Message Passing Neural Network (MPNN)}:  A generalization of the GCN, this model allows for more flexible aggregation of neighbor features, and has additional nonlinear feature transformations possible. \cite{gilmer2017neural}
    \item \textbf{Graph Isomorphism Network (GIN)}: A GNN designed specifically to maximize the expressiveness of the model, with respect to a classic graph isomorphim test. \cite{gin}
    \item \textbf{Multi-Head Attention (Graph Transformer)}: This GNN adapts transformer style attention, allowing it to learn different ways of passing messages (based on the attention mask). \cite{dwivedi2021generalization}
    \item \textbf{Heterogeneous Graph Transformer (HGT)}:  Another adoption of transformer style attention (we note that it can be applied to non-heterogeneous graphs as well). \cite{hu2020heterogeneous}
    \item \textbf{Linear Aggregation}  In addition to the popular encoders from the literature, we also evaluated a family of models which use linear aggregation of features, as this has been shown to work surprisingly well on a number of tasks \cite{bojchevski2020scaling}.
    \begin{itemize}
        \item \textbf{Node Set}:  This model simply pools all the node features in the graph together.
        \item \textbf{Edge Set}:  This model simply pools all the edge features together (edge features are defined as the concatenation of its two nodes' features).
    \end{itemize}

\end{itemize}

\textbf{Setting}:
The experimental setup is similar to the experiment in Section \ref{sec:exp1_graphtoken}. Again, GraphQA$_\text{Train}$ performance was used for model selection, and we report the corresponding model's results on GraphQA$_\text{Test}$.

\textbf{Result}: \Cref{table:encoder_choice} shows the results for each model on GraphQA$_\text{Test}$.
In general, we see that there is no one model that consistently dominates across graph encoding tasks.
Instead, we see that different graph encoder architectures have strengths and weaknesses advantages.

There is one notable pattern however, is that the simple linear GNN models perform quite strongly at their respective counting tasks (i.e.\ NodeSet does well at \task{node count}, EdgeSet does well at \task{edge count}).  However models with non-linear effects are still capable on these tasks (\eg{} MHA does well at \task{node count}, and MPNN does well on \task{edge count}). 

\subsubsection{Choice: GNN Features}
\label{subsec:gnn_features}   
Recently, positional node encodings~\cite{wang2022equivariant,dwivedi2023benchmarking,lim2023sign} were proposed to enhance the expressivity of GNNs.
On the other hand, in molecular modeling it has been shown recently that non-equivariant encoders can match or exceed quality of equivariant ones~\cite{wang2023generating}.
This raises a more general question: do GNNs need to be equivariant in order to generalize, especially with extremely powerful decoders, such as LLMs?

We investigate this question by testing the graph reasoning capability of \ourmodel{} with three distinct node featurization settings:
\begin{itemize}[topsep=0pt,itemsep=-1ex,partopsep=1ex,parsep=1ex]
    \item \textbf{LPE}: Laplacian positional encodings using the normalized Laplacian matrix, as in~\cite{dwivedi2023benchmarking}.
    \item \textbf{IDX}: unique identity encoding designed to break the GNN equivariance.
    \item \textbf{LPE+IDX}: a concatenation of the above two strategies.
\end{itemize}

\textbf{Setting.}  The experimental setup is similar to \ref{sec:encoder_design}.
Node features of dimensionality $d=4$ are evaluated for LPE and IDX featurization.  Models using LPE+IDX contains node features of size $d=8$.

\begin{figure*}[t!]
\centering
     \begin{subfigure}[b]{0.33\linewidth}
         \centering
         \includegraphics[width=\linewidth]{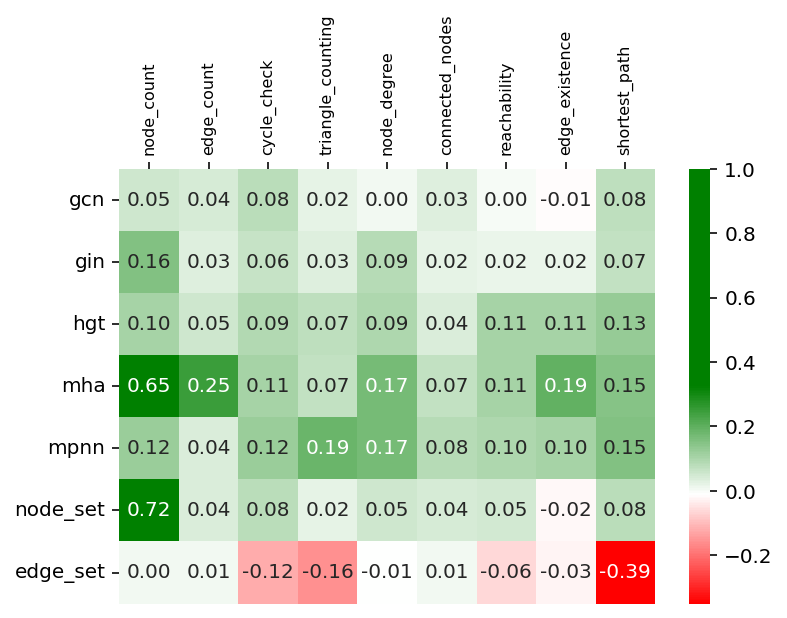}
         \caption{Spectral Features (LPE)}
         \label{fig:node_features-lpe}
     \end{subfigure}
\begin{subfigure}[b]{0.33\linewidth}
         \centering
         \includegraphics[width=\linewidth]{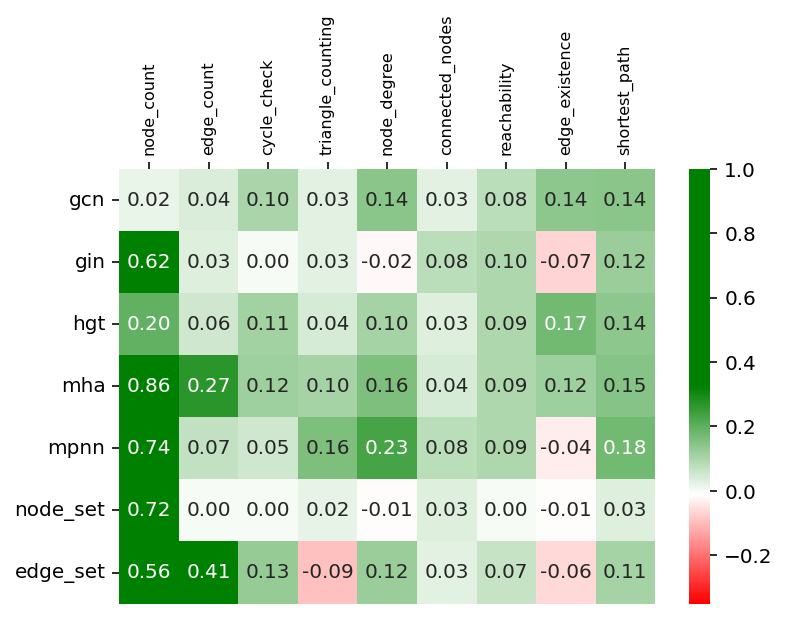}
         \caption{Learned Features (IDX)}
         \label{fig:node_features-idx}
     \end{subfigure}
\begin{subfigure}[b]{0.33\linewidth}
         \centering
         \includegraphics[width=\linewidth]{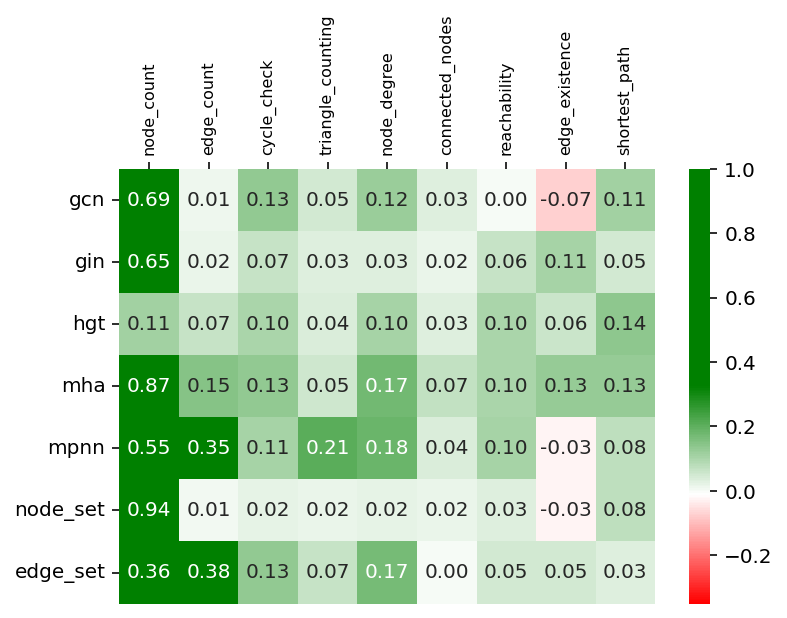}
         \caption{Learned and Spectral Features (LPE+IDX)}
         \label{fig:node_features-lpe_idx}
     \end{subfigure}
    \caption{Effect of varying node features used in the graph encoder.  Results shown are performance difference from the \textsc{Soft Prompt} baseline on GraphQA$_{\text{Test}}$.  We see that breaking equivariance via learned features (Fig. \ref{fig:node_features-idx}) generally improve the model performance, but the combination of learned and spectral features (Fig. \ref{fig:node_features-lpe_idx}) proves uniquely powerful for some encoders.}
    \label{fig:feature_choice}
\end{figure*}

\textbf{Result}.  
The outcome of this experiment are show in Figure \ref{fig:feature_choice}, where we see the difference of all models from the \textsc{SoftPrompt}  baseline \cite{lester2021power} when evaluted on GraphQA$_{\text{Test}}$. The core result is that learned positional embeddings (Fig. \ref{fig:node_features-idx})  generally outperform Laplacian position embeddings (Fig \ref{fig:node_features-lpe}) for most encoders and most tasks.
These results show that breaking equivariance surprisingly adds additional capabilities for graph reasoning when powerful decoders are present.  Some additional observations include:
\begin{itemize}
    \item 
\textit{Counting Tasks}.  Learned features seem to provide essential lift for basic counting tasks (\task{node count}, \task{edge count}, and \task{node degree}) in many encoders.
    \item \textit{Combination}.  In some very interesting cases of task and encoder, the combination of both types of features greatly improved model performance (Fig. \ref{fig:node_features-lpe_idx}).  For example, GCN and NodeSet significantly improved at the \task{node count} task.
    \item \textit{Linear models}.  NodeSet (an encoder which does not use the graph edges) generally benefited from spectral features as they added previously unseen information about the graph structure.
\end{itemize}

\begin{table}[t!]
\centering
\newcolumntype{C}{>{\raggedleft\arraybackslash}X}
\newcolumntype{S}{>{\raggedright\arraybackslash}X}
\footnotesize
\caption{Total number of parameters in the graph encoder.\label{tbl:n_params}}
\begin{tabularx}{\linewidth}{p{2cm}CC}
\toprule
{} &  Body  &  Head\\
\midrule
GCN      &            17,152 &         $1.1 \times 10^7$ \\
GIN      &            17,152 &         $1.1 \times 10^7$ \\
MPNN  &            83,968 &         $1.1 \times 10^7$ \\
HGT      &           198,788 &         $1.1 \times 10^7$ \\
MHA      &           101,376 &         $1.1 \times 10^7$ \\
\midrule
Node Set      &                0 &           $4.1 \times 10^5$ \\
Edge Set &                0 &           $7.4 \times 10^5$ \\
\bottomrule
\end{tabularx}
\end{table} 
\subsubsection{Parameter Usage in \ourmodel{}}

\textbf{Setting}: We consider the graph convolution evaluation from Section \ref{sec:exp2_graph_conv}, using LPE features with dimensionality $d=4$. 
The graph encoders have a latent space of size $128$.
We then project this into a prompt embedding with approximately $80,000$ parameters in \ourmodel{} .

\textbf{Results}:
\Cref{tbl:n_params} shows the number of parameters used in the graph encoder.
Here `body' refers to the number of parameters in the graph encoder itself, and `Head'  refers to the parameters in the transformation layer to the higher-dimensional LLM token space.

Its also insightful to consider the number of parameters used in each of the models.
Table~\ref{tbl:n_params} specifies total number of parameters used by each GNN architecture.
We note that this size is dominated by the total number of parameters in the projection layer to the token space (roughly 11 million).
Out of all non-linear architectures, attention-based ones (MHA and HGT) add the most encoder-based parameters.
In general, the size of our graph encoder models varies from 17k to 199k parameters. This is \textit{significantly smaller} than typical LLMs, which currently often contain tens or hundreds of billions of parameters. For example, the open-source LLama2 language model scales from 7 billion to 70 billion parameters~\cite{touvron2023llama}.
Meanwhile the closed source PaLM 1 model contains 540 billion parameters~\cite{chowdhery2022palm}.
In light of this, we can see that \ourmodel{} is highly parameter-efficient, and significantly improves the graph reasoning capability of a LLM while barely adding any parameters at all.

\section{Discussion}
So far we have examined the performance benefits of \ourmodel{}, and the design choices necessary when building a graph encoder.
However several questions remain: (1) What exactly are the encoders doing, and (2) does it generalize?
In this section we seek to provide some insight (if not answers) to these questions, and lay the foundations for future work.

\subsection{Graph Encoder Analysis}

\begin{figure}[!t]
    \centering
\includegraphics[width=.7\columnwidth]{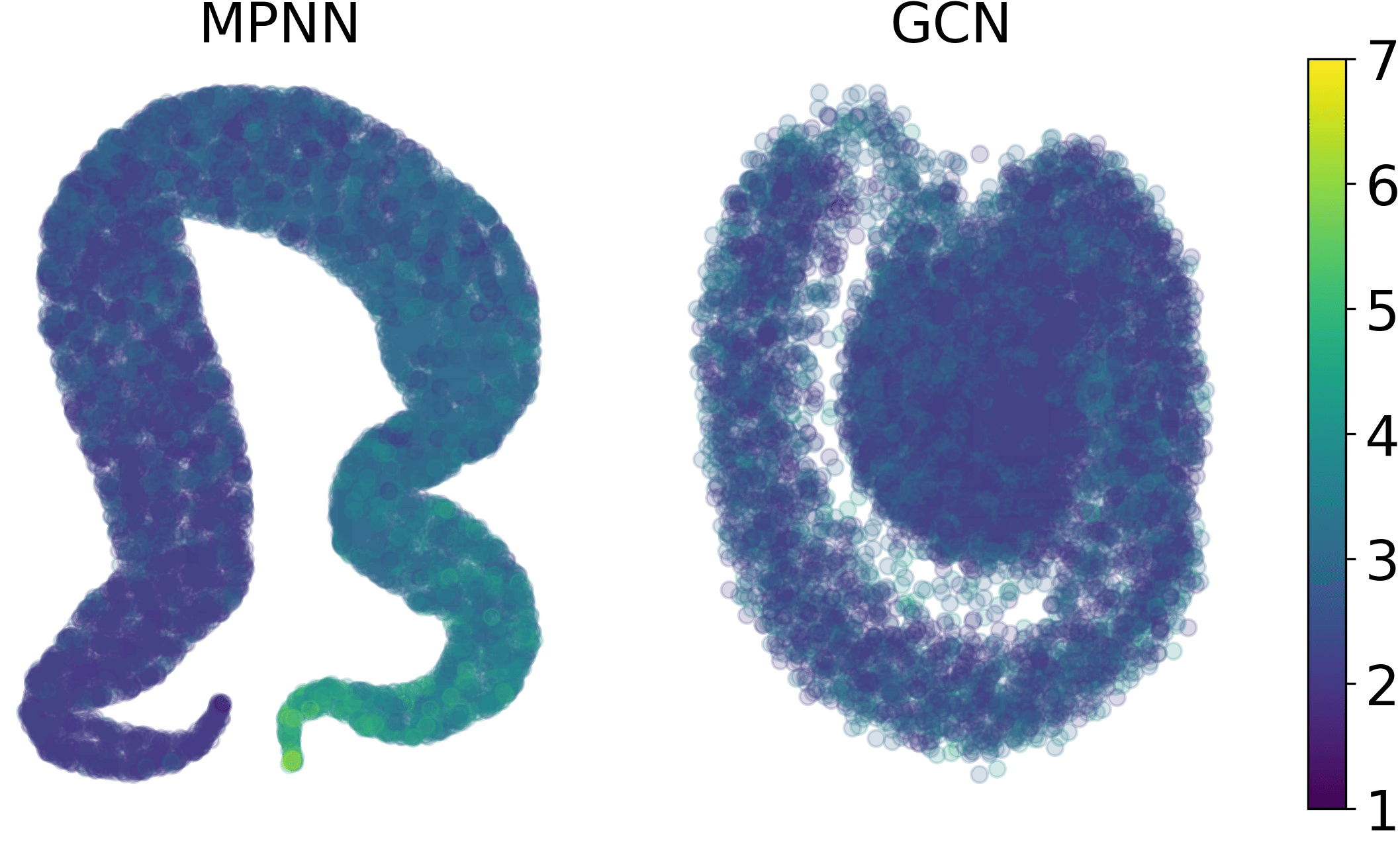}
    \caption{UMAP~\cite{mcinnes2018umap} projection of \ourmodel{} embeddings produced by two different encoders, colored by the diameter of a graph. We plot all 8-node graphs.}
    \label{fig:umap_embeddings}
\end{figure}

\begin{table*}[t]
\centering
\caption{Predicting \textbf{bipartiteness} using graph encoders trained for different tasks, measured on all graphs with 8 nodes. 
Observe that graph encoders trained on \texttt{cycle check} and \texttt{triangle counting} generalize well to bipartiteness detection.
\label{tbl:knn_bipartite_s8}}
\footnotesize
\resizebox{\textwidth}{!}{\setlength{\tabcolsep}{4pt}
\begin{tabular}{ll|cccc|cc|ccc} 
\toprule
 & & \multicolumn{4}{c}{\textbf{Graph Tasks}} & \multicolumn{2}{c}{\textbf{Node Tasks}} & \multicolumn{3}{c}{\textbf{Edge Tasks}} \\
\cmidrule(lr){3-6} \cmidrule(lr){7-8} \cmidrule(lr){9-11}
& \textbf{Method} & \textbf{Node count} & \textbf{Edge count} & \textbf{Cycle check} & \textbf{Triangle counting} & \textbf{Node degree} & \textbf{Connected nodes} & \textbf{Reachability} & \textbf{Edge existence} & \textbf{Shortest path} \\
\midrule
\scriptsize{\multirow{5}{*}{\begin{sideways}Non-linear\end{sideways}}} & GCN & 53.82 & 53.28 & 55.46 & 50.00 & 50.00 & 54.64 & 50.00 & 48.48 & 51.60 \\
& GIN & 51.09 & 53.27 & 52.74 & 51.91 & 53.26 & 53.57 & 51.36 & 52.17 & 52.18 \\
& MPNN & \textbf{68.01} & \textbf{71.34} & 56.82 & 76.82 & \textbf{60.13} & 60.95 & 61.77 & 62.58 & 54.37 \\
& HGT & 50.00 & 54.35 & 68.53 & \textbf{95.03} & 50.27 & 59.81 & \textbf{68.85} & \textbf{74.58} & 50.00 \\
& MHA & 50.27 & 66.39 & \textbf{87.00} & 72.14 & 58.74 & \textbf{66.38} & 51.63 & 54.12 & \textbf{64.45} \\
\midrule
\scriptsize{\multirow{2}{*}{\begin{sideways}Linear\end{sideways}}} & Node Set & 56.55 & 57.38 & 56.30 & 55.74 & 56.29 & 56.28 & 55.73 & 57.93 & 56.56 \\
& Edge Set & 50.82 & 50.82 & 50.82 & 50.55 & 50.54 & 50.54 & 50.82 & 50.82 & 50.54 \\
\bottomrule
\end{tabular}
}
\end{table*} 
This section studies the properties learned by \ourmodel{}'s graph encoders by directly examining the representations they produce.
We study both the in-distribution and out-of-distribution properties of these encoders.
We consider 9 tasks in total: total number of edges; maximum node degree; graph diameter; number of triangles; average local clustering coefficient; largest core number; average shortest path length; testing planarity; testing bipartiteness.

One benefit of studying graphs is data availability: for small-enough graphs, we can generate all possible graphs \emph{exhaustively} using \texttt{geng}~\cite{mckay1981practical}.
The evaluation goes as follows:  First, we train an encoder on a task from GraphQA (e.g.\ \texttt{cycle check}).
Then, to evaluate the cross-task generalizability of the different encoders 
we train a kNN classifier (or regressor) with $k=5$ on the representations of (i) an exhaustive set of connected graphs with 8 nodes (called \texttt{graph8c} in~\citet{balcilar2021breaking}) and (ii) an exhaustive set of tree graphs with 15 nodes. 
We note that because we are generating a large set of graphs (e.g. there are 11117 graphs of size 8) and only trained on GraphQA$_{\text{Train}}$ (1000 instances), the vast majority of the graphs we are using here are unseen.
As an illustration, a UMAP \cite{mcinnes2018umap} visualization of the embeddings for all 8 node graphs using two GNN encoders is presented in Figure \ref{fig:umap_embeddings}.

\textbf{Results}. Since we present a lot of experiments and it's hard to cover them all, we focus here on the task of predicting whether a graph is bipartite and outsource the rest to the Appendix.
From the basic graph theory we know that, if there is a triangle or an odd cycle in a graph, it can not be bipartite.
Therefore, we expect \task{triangle counting} and \task{cycle check} training objectives to perform well on this task.
In Table \ref{tbl:knn_bipartite_s8} we can see that this is precisely what happens, with attention-based methods taking the lead.
This is an interesting example of \textit{generalization} from the graph encoders to a new task.

Overall, there is a significant performance gap between different graph encoders, MPNN and attention-based ones being generally the best.
We observe significant correlations in performance of in-distribution learning -- for instance, \ourmodel{} trained on \task{edge count} performs the best on \task{edge count} prediction.
What is interesting is that \task{node count} performs comparably here.
This suggests that graph encoders learn some universal features that are applicable to many different downstream tasks.

\subsection{Future Work}

This work opens up an exciting new avenue of exploration for reasoning with structured data and LLMs. Some potential
avenues that we consider particularly exciting include:

\begin{itemize}
\item This work just considers existing convolutions and measures their effectiveness.  An obvious and essential next step is designing graph convolutions that best support LLMs in various graph reasoning tasks.
    \item Evaluating the usefulness of this approach for factual grounding. Can we improve the ability of an LLM to answer questions about the data using prompting over knowledge graphs? Could an LLM answer novel questions about a molecule given a GNN-produced representation of it?
    \item \ourmodel{} improves performance with broken equivariance. Can this result inform other problems with very strong decoder models?
    \item This work examines how a GNN can be used to an enhance LLMs, but what about the reverse? Can we use an LLM to interrogate a GNN to better explain its results or provide higher quality answers?
\end{itemize}

\section{Conclusions}

In this work we have studied the structured data encoding problem for LLMs.  
Our novel method, \ourmodel{}, learns a graph embedding function through the gradients provided by a \textit{frozen} LLM.
\ourmodel{} is especially well suited for projecting structured data into latent `prompt space'.
It is a parameter-efficient method as it requires only training the graph encoder and does not update LLM parameters.
Our extensive experimental analysis across 9 graph reasoning tasks shows that \ourmodel{} greatly improves graph reasoning in LLMs -- we observe up to a 73\% improvement on the GraphQA benchmark.

There is still much to do!  We believe that our approach is fundamental for adapting new structured data sources to LLMs (which are expensive and time consuming to train), and presents a very attractive way of improving fundamental problems in LLMs, including \textit{hallucinations}, \textit{factuality}, and \textit{freshness}.

\clearpage
\section*{Acknowledgements}
We thank Oleksandr Ferludin, Johannes Gasteiger, Silvio Lattanzi, Vahab Mirrokni and Jan Pfeifer for discussions about the work.

\bibliography{references}
\bibliographystyle{preprint}

\newpage
\appendix
\onecolumn
\section{Appendix}
\subsection{Graph Encoders}

\textbf{Notation.}  We briefly describe the notation we will use. The graph $G=(V,E)$ contains the set of $V$ nodes and $E$ edges. While we will only discuss simple graphs, everything discussed can be extended to heterogeneous graphs w.l.o.g.~\cite{battaglia2018relational,ferludin2023tfgnn}.

Using the notation of \citet{ferludin2023tfgnn}, a GNN has two primary operations. First, a next state function ($\textsc{NextState}$) which computes the hidden state $\mathbf{h}_v$ of a node (or edge,  $\mathbf{m}_{(u,v)}$) given information from its neighbors and its previous state, and an aggregation function ($\textsc{EdgePool}$) which pools information for a node's immediate neighborhood into a fixed size representation.
More formally, we can say that the next state of a node is: 
\begin{equation*}
  \label{eq:next-node-state}
    \mathbf{h}_v^{(i+1)} = \textsc{NextState}_V^{(i+1)}(
        \mathbf{h}_v^{(i)},
        \overline{\mathbf{m}}_{v}^{(i+1)}).
\end{equation*}
Then the pooled messages $\overline{\mathbf{m}}_{v}^{(i+1)}$ are defined as follows:
\begin{equation*}
  \label{eq:mbar-two-steps}
  \begin{split}
      \mathbf{m}_{(u,v)}^{(i+1)} &= \textsc{NextState}_{E}^{(i+1)}(
      \mathbf{h}_u^{(i)},
      \mathbf{h}_v^{(i)},
      \mathbf{m}_{(u,v)}^{(i)}),\\
    \overline{\mathbf{m}}_{v}^{(i+1)} &= \textsc{EdgePool}^{(i+1)}(
      \mathbf{h}_v^{(i)},
      \{ \mathbf{m}_{(u,v)}^{(i+1)} \mid
         u \in \mathcal{N}(v)\}).
  \end{split}
\end{equation*}
Different realizations of the $\textsc{NextState}$ and $\textsc{EdgePool}$ functions can implement a wide variety of GNN operations. This can include powerful models which use Transformer style attention instead of the provided graph edges \cite{dwivedi2021generalization}.

\begin{figure*}[h!]
     \centering
     \begin{subfigure}[b]{0.3\linewidth}
         \centering
         \includegraphics[width=\linewidth]{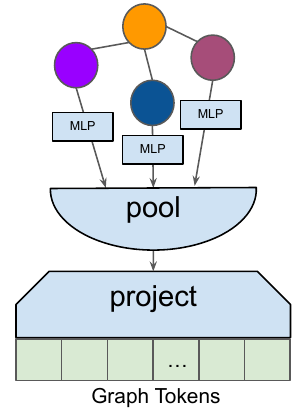}
         \caption{Node Set architecture}
         \label{fig:node_set_arch}
     \end{subfigure}
     \begin{subfigure}[b]{0.3\linewidth}
         \centering
         \includegraphics[width=\linewidth]{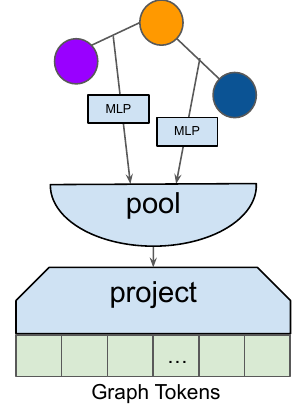}
         \caption{Edge Set architecture}
         \label{fig:edge_set_arch}
     \end{subfigure}
    \caption{Figurative illustrations of set-based GNN architectures employed in the paper. We pool representations from either nodes or edges, transform them via an MLP with shared weights, pool, and project to the \ourmodel{} space.}
    \label{fig:set_architectures}
\end{figure*}

The architecture of NodeSet and EdgeSet is shown in Figure \ref{fig:set_architectures}.
Other GNN models have graph convolutions before node/edge states are read out.

\subsection{Additional experiments}

We present additional results for graph encoder analysis.
Tables~\ref{tbl:knn_clustering_s8}--\ref{tbl:knn_sp_length_t15} present additional results on more graph properties, as well as experiments on tree-structured graphs of size 15.
In general, complete graph populations demonstrate significantly better performance than trees -- we can attribute that to the fact that \ourmodel{} was trained on diverse sets of data, and trees are somewhat out-of-distribution.
Nevertheless, for all considered cases the best overall encoder model achieved better results than na\"ive set encodings.

\begin{table*}[htbp]
\centering
\caption{Average local clustering coefficient MSE measured on all connected graphs with 8 nodes. We highlight the best performance per training task in columns.\label{tbl:knn_clustering_s8}}
\footnotesize
\resizebox{\textwidth}{!}{\setlength{\tabcolsep}{4pt}
\begin{tabular}{ll|cccc|cc|ccc} 
\toprule
 & & \multicolumn{4}{c}{\textbf{Graph Tasks}} & \multicolumn{2}{c}{\textbf{Node Tasks}} & \multicolumn{3}{c}{\textbf{Edge Tasks}} \\
\cmidrule(lr){3-6} \cmidrule(lr){7-8} \cmidrule(lr){9-11}
& \textbf{Method} & \textbf{Node count} & \textbf{Edge count} & \textbf{Cycle check} & \textbf{Triangle counting} & \textbf{Node degree} & \textbf{Connected nodes} & \textbf{Reachability} & \textbf{Edge existence} & \textbf{Shortest path} \\
\midrule
\scriptsize{\multirow{5}{*}{\begin{sideways}Non-linear\end{sideways}}} & GCN & 1.62 & 1.67 & 2.12 & 4.49 & 4.49 & 1.73 & 4.49 & 16.57 & 3.75 \\
& GIN & 2.18 & 2.29 & 2.45 & 2.60 & 2.44 & 2.31 & 3.73 & 2.88 & 3.37 \\
& MPNN & \textbf{1.03} & \textbf{0.95} & 1.38 & \textbf{0.81} & \textbf{1.50} & 1.34 & \textbf{1.68} & 1.87 & 1.47 \\
& HGT & 2.63 & 2.25 & 2.08 & 1.23 & 2.49 & 2.17 & 1.90 & 1.62 & 2.52 \\
& MHA & 2.69 & 1.01 & \textbf{1.23} & 0.96 & 1.56 & \textbf{1.25} & 2.08 & \textbf{1.59} & \textbf{1.29} \\
\midrule
\scriptsize{\multirow{2}{*}{\begin{sideways}Linear\end{sideways}}} & Node Set & 2.59 & 2.56 & 2.59 & 2.59 & 2.58 & 2.60 & 2.58 & 2.58 & 2.56 \\
& Edge Set & 2.22 & 2.22 & 2.22 & 2.22 & 2.24 & 2.23 & 2.22 & 2.22 & 2.23 \\
\bottomrule
\end{tabular}
}
\end{table*} \begin{table*}[htbp]
\centering
\caption{Degree accuracy on all connected graphs with 8 nodes. We highlight the best performance per training task in columns.}\label{tbl:knn_degree_s8}
\footnotesize
\resizebox{\textwidth}{!}{\setlength{\tabcolsep}{4pt}
\begin{tabular}{ll|cccc|cc|ccc} 
\toprule
 & & \multicolumn{4}{c}{\textbf{Graph Tasks}} & \multicolumn{2}{c}{\textbf{Node Tasks}} & \multicolumn{3}{c}{\textbf{Edge Tasks}} \\
\cmidrule(lr){3-6} \cmidrule(lr){7-8} \cmidrule(lr){9-11}
& \textbf{Method} & \textbf{Node count} & \textbf{Edge count} & \textbf{Cycle check} & \textbf{Triangle counting} & \textbf{Node degree} & \textbf{Connected nodes} & \textbf{Reachability} & \textbf{Edge existence} & \textbf{Shortest path} \\
\midrule
\scriptsize{\multirow{5}{*}{\begin{sideways}Non-linear\end{sideways}}} & GCN & 57.46 & 56.65 & 52.46 & 40.09 & 40.09 & 57.42 & 40.09 & 15.73 & 40.26 \\
& GIN & 56.86 & 56.30 & 54.55 & 48.75 & 55.59 & 57.56 & 40.14 & 50.81 & 44.83 \\
& MPNN & \textbf{69.45} & \textbf{69.60} & \textbf{67.19} & \textbf{71.84} & \textbf{64.56} & \textbf{67.62} & 61.37 & 58.66 & 63.18 \\
& HGT & 55.20 & 55.70 & 56.54 & 60.17 & 56.62 & 57.65 & 58.02 & 59.06 & 55.46 \\
& MHA & 54.86 & 64.33 & 62.86 & 65.63 & 61.67 & 63.22 & 56.98 & 61.60 & \textbf{63.97} \\
\midrule
\scriptsize{\multirow{2}{*}{\begin{sideways}Linear\end{sideways}}} & Node Set & 54.66 & 54.91 & 54.98 & 55.06 & 54.78 & 54.64 & 54.50 & 54.94 & 54.72 \\
& Edge Set & 63.48 & 63.37 & 63.07 & 63.55 & 63.08 & 63.37 & \textbf{63.47} & \textbf{63.06} & 63.44 \\
\bottomrule
\end{tabular}
}
\end{table*} \begin{table*}[htbp]
\centering
\caption{Diameter Accuracy on all connected graphs with 8 nodes. We highlight the best performance per training task in columns.}\label{tbl:knn_diameter_s8}
\footnotesize
\resizebox{\textwidth}{!}{\setlength{\tabcolsep}{4pt}
\begin{tabular}{ll|cccc|cc|ccc} 
\toprule
 & & \multicolumn{4}{c}{\textbf{Graph Tasks}} & \multicolumn{2}{c}{\textbf{Node Tasks}} & \multicolumn{3}{c}{\textbf{Edge Tasks}} \\
\cmidrule(lr){3-6} \cmidrule(lr){7-8} \cmidrule(lr){9-11}
& \textbf{Method} & \textbf{Node count} & \textbf{Edge count} & \textbf{Cycle check} & \textbf{Triangle counting} & \textbf{Node degree} & \textbf{Connected nodes} & \textbf{Reachability} & \textbf{Edge existence} & \textbf{Shortest path} \\
\midrule
\scriptsize{\multirow{5}{*}{\begin{sideways}Non-linear\end{sideways}}} & GCN & 66.86 & 67.81 & 66.70 & 37.37 & 37.37 & 68.91 & 37.37 & 52.13 & 55.13 \\
& GIN & 66.06 & 64.87 & 63.97 & 61.09 & 64.98 & 66.43 & 37.80 & 60.65 & 54.82 \\
& MPNN & \textbf{76.92} & \textbf{76.86} & 73.63 & \textbf{78.33} & \textbf{74.78} & \textbf{77.18} & \textbf{74.42} & \textbf{69.56} & \textbf{76.23} \\
& HGT & 63.97 & 65.24 & 66.88 & 70.45 & 65.30 & 68.45 & 69.64 & 68.97 & 66.04 \\
& MHA & 63.76 & 74.17 & \textbf{76.00} & 74.03 & 73.50 & 74.71 & 68.45 & 69.32 & 72.95 \\
\midrule
\scriptsize{\multirow{2}{*}{\begin{sideways}Linear\end{sideways}}} & Node Set & 67.28 & 67.24 & 67.01 & 66.97 & 66.81 & 67.19 & 67.09 & 66.87 & 66.79 \\
& Edge Set & 66.99 & 66.51 & 66.63 & 66.83 & 66.65 & 67.02 & 66.60 & 66.93 & 66.90 \\
\bottomrule
\end{tabular}
}
\end{table*} \begin{table*}[htbp]
\centering
\caption{k-Core Accuracy on all connected graphs with 8 nodes. We highlight the best performance per training task in columns.}\label{tbl:knn_kcore_s8}
\footnotesize
\resizebox{\textwidth}{!}{\setlength{\tabcolsep}{4pt}
\begin{tabular}{ll|cccc|cc|ccc} 
\toprule
 & & \multicolumn{4}{c}{\textbf{Graph Tasks}} & \multicolumn{2}{c}{\textbf{Node Tasks}} & \multicolumn{3}{c}{\textbf{Edge Tasks}} \\
\cmidrule(lr){3-6} \cmidrule(lr){7-8} \cmidrule(lr){9-11}
& \textbf{Method} & \textbf{Node count} & \textbf{Edge count} & \textbf{Cycle check} & \textbf{Triangle counting} & \textbf{Node degree} & \textbf{Connected nodes} & \textbf{Reachability} & \textbf{Edge existence} & \textbf{Shortest path} \\
\midrule
\scriptsize{\multirow{5}{*}{\begin{sideways}Non-linear\end{sideways}}} & GCN & 69.49 & 69.15 & 66.61 & 58.33 & 58.33 & 69.16 & 58.33 & 25.18 & 61.55 \\
& GIN & 68.03 & 65.98 & 64.85 & 62.67 & 66.74 & 67.84 & 58.84 & 63.34 & 59.08 \\
& MPNN & \textbf{87.42} & \textbf{87.54} & \textbf{81.81} & \textbf{88.63} & \textbf{80.30} & \textbf{83.48} & \textbf{80.08} & 71.01 & \textbf{82.05} \\
& HGT & 63.92 & 65.29 & 67.00 & 70.01 & 65.44 & 67.32 & 68.35 & 70.08 & 65.13 \\
& MHA & 64.30 & 80.80 & 73.49 & 80.81 & 76.98 & 78.83 & 69.43 & \textbf{74.21} & 75.92 \\
\midrule
\scriptsize{\multirow{2}{*}{\begin{sideways}Linear\end{sideways}}} & Node Set & 68.23 & 68.74 & 68.50 & 68.71 & 68.07 & 67.99 & 68.85 & 68.17 & 68.70 \\
& Edge Set & 66.30 & 65.78 & 65.58 & 66.15 & 65.76 & 65.91 & 65.94 & 65.77 & 65.71 \\
\bottomrule
\end{tabular}
}
\end{table*} \begin{table*}[htbp]
\centering
\caption{\#edges Accuracy on all connected graphs with 8 nodes. We highlight the best performance per training task in columns.}\label{tbl:knn_nedges_s8}
\footnotesize
\resizebox{\textwidth}{!}{\setlength{\tabcolsep}{4pt}
\begin{tabular}{ll|cccc|cc|ccc} 
\toprule
 & & \multicolumn{4}{c}{\textbf{Graph Tasks}} & \multicolumn{2}{c}{\textbf{Node Tasks}} & \multicolumn{3}{c}{\textbf{Edge Tasks}} \\
\cmidrule(lr){3-6} \cmidrule(lr){7-8} \cmidrule(lr){9-11}
& \textbf{Method} & \textbf{Node count} & \textbf{Edge count} & \textbf{Cycle check} & \textbf{Triangle counting} & \textbf{Node degree} & \textbf{Connected nodes} & \textbf{Reachability} & \textbf{Edge existence} & \textbf{Shortest path} \\
\midrule
\scriptsize{\multirow{5}{*}{\begin{sideways}Non-linear\end{sideways}}} & GCN & 38.91 & 39.19 & 35.94 & 11.60 & 11.60 & 40.24 & 11.60 & 2.19 & 14.58 \\
& GIN & 38.13 & 37.33 & 36.57 & 31.66 & 37.74 & 38.34 & 11.88 & 31.45 & 25.92 \\
& MPNN & \textbf{86.58} & \textbf{86.72} & \textbf{53.15} & \textbf{84.56} & \textbf{52.12} & \textbf{66.01} & \textbf{50.70} & \textbf{41.96} & \textbf{59.95} \\
& HGT & 35.63 & 37.45 & 38.23 & 40.39 & 37.14 & 37.80 & 39.68 & 39.74 & 36.86 \\
& MHA & 35.85 & 55.32 & 45.04 & 53.52 & 47.89 & 49.44 & 39.69 & 42.84 & 46.17 \\
\midrule
\scriptsize{\multirow{2}{*}{\begin{sideways}Linear\end{sideways}}} & Node Set & 40.06 & 40.14 & 39.40 & 40.15 & 39.97 & 39.72 & 39.88 & 39.79 & 39.89 \\
& Edge Set & 37.93 & 38.11 & 38.05 & 37.92 & 38.05 & 37.67 & 37.64 & 37.82 & 37.91 \\
\bottomrule
\end{tabular}
}
\end{table*} \begin{table*}[htbp]
\centering
\caption{Planarity AUC on all connected graphs with 8 nodes. We highlight the best performance per training task in columns.}\label{tbl:knn_planar_s8}
\footnotesize
\resizebox{\textwidth}{!}{\setlength{\tabcolsep}{4pt}
\begin{tabular}{ll|cccc|cc|ccc} 
\toprule
 & & \multicolumn{4}{c}{\textbf{Graph Tasks}} & \multicolumn{2}{c}{\textbf{Node Tasks}} & \multicolumn{3}{c}{\textbf{Edge Tasks}} \\
\cmidrule(lr){3-6} \cmidrule(lr){7-8} \cmidrule(lr){9-11}
& \textbf{Method} & \textbf{Node count} & \textbf{Edge count} & \textbf{Cycle check} & \textbf{Triangle counting} & \textbf{Node degree} & \textbf{Connected nodes} & \textbf{Reachability} & \textbf{Edge existence} & \textbf{Shortest path} \\
\midrule
\scriptsize{\multirow{5}{*}{\begin{sideways}Non-linear\end{sideways}}} & GCN & 74.18 & 73.76 & 72.61 & 50.00 & 50.00 & 74.74 & 50.00 & 50.00 & 49.44 \\
& GIN & 77.35 & 73.00 & 72.06 & 69.37 & 74.86 & 75.85 & 50.73 & 68.97 & 61.58 \\
& MPNN & \textbf{86.14} & \textbf{86.52} & \textbf{84.16} & \textbf{86.64} & \textbf{83.74} & \textbf{85.17} & \textbf{84.32} & 77.84 & \textbf{85.55} \\
& HGT & 69.24 & 71.41 & 71.02 & 74.07 & 71.47 & 72.20 & 72.20 & 73.59 & 71.55 \\
& MHA & 69.96 & 80.87 & 78.35 & 80.46 & 81.53 & 81.21 & 74.98 & \textbf{78.29} & 80.58 \\
\midrule
\scriptsize{\multirow{2}{*}{\begin{sideways}Linear\end{sideways}}} & Node Set & 78.41 & 78.76 & 78.86 & 78.82 & 78.18 & 78.54 & 78.72 & 78.76 & 78.78 \\
& Edge Set & 72.17 & 71.64 & 72.06 & 72.20 & 71.93 & 72.11 & 72.01 & 72.27 & 72.01 \\
\bottomrule
\end{tabular}
}
\end{table*} \begin{table*}[htbp]
\centering
\caption{Shortest path MSE on all connected graphs with 8 nodes. We highlight the best performance per training task in columns.}\label{tbl:knn_sp_length_s8}
\footnotesize
\resizebox{\textwidth}{!}{\setlength{\tabcolsep}{4pt}
\begin{tabular}{ll|cccc|cc|ccc} 
\toprule
 & & \multicolumn{4}{c}{\textbf{Graph Tasks}} & \multicolumn{2}{c}{\textbf{Node Tasks}} & \multicolumn{3}{c}{\textbf{Edge Tasks}} \\
\cmidrule(lr){3-6} \cmidrule(lr){7-8} \cmidrule(lr){9-11}
& \textbf{Method} & \textbf{Node count} & \textbf{Edge count} & \textbf{Cycle check} & \textbf{Triangle counting} & \textbf{Node degree} & \textbf{Connected nodes} & \textbf{Reachability} & \textbf{Edge existence} & \textbf{Shortest path} \\
\midrule
\scriptsize{\multirow{5}{*}{\begin{sideways}Non-linear\end{sideways}}} & GCN & 2.27 & 2.24 & 2.31 & 6.07 & 6.07 & 2.06 & 6.07 & 11.09 & 3.75 \\
& GIN & 2.57 & 2.77 & 2.83 & 2.93 & 2.52 & 2.54 & 4.84 & 3.09 & 3.61 \\
& MPNN & \textbf{0.29} & \textbf{0.29} & \textbf{0.76} & \textbf{0.31} & \textbf{0.71} & \textbf{0.49} & \textbf{0.75} & 1.58 & \textbf{0.51} \\
& HGT & 3.03 & 2.64 & 2.27 & 1.60 & 2.60 & 2.14 & 1.80 & 1.95 & 2.81 \\
& MHA & 3.04 & 0.71 & 0.95 & 0.78 & 1.01 & 0.74 & 1.74 & \textbf{1.55} & 1.05 \\
\midrule
\scriptsize{\multirow{2}{*}{\begin{sideways}Linear\end{sideways}}} & Node Set & 2.35 & 2.35 & 2.35 & 2.36 & 2.36 & 2.35 & 2.34 & 2.36 & 2.34 \\
& Edge Set & 2.99 & 2.99 & 2.99 & 2.99 & 2.97 & 2.97 & 2.99 & 2.99 & 2.99 \\
\bottomrule
\end{tabular}
}
\end{table*} \begin{table*}[htbp]
\centering
\caption{\# of triangles MSE on all connected graphs with 8 nodes. We highlight the best performance per training task in columns.}\label{tbl:knn_triangles_s8}
\footnotesize
\resizebox{\textwidth}{!}{\setlength{\tabcolsep}{4pt}
\begin{tabular}{ll|cccc|cc|ccc} 
\toprule
 & & \multicolumn{4}{c}{\textbf{Graph Tasks}} & \multicolumn{2}{c}{\textbf{Node Tasks}} & \multicolumn{3}{c}{\textbf{Edge Tasks}} \\
\cmidrule(lr){3-6} \cmidrule(lr){7-8} \cmidrule(lr){9-11}
& \textbf{Method} & \textbf{Node count} & \textbf{Edge count} & \textbf{Cycle check} & \textbf{Triangle counting} & \textbf{Node degree} & \textbf{Connected nodes} & \textbf{Reachability} & \textbf{Edge existence} & \textbf{Shortest path} \\
\midrule
\scriptsize{\multirow{5}{*}{\begin{sideways}Non-linear\end{sideways}}} & GCN & 132.94 & 129.03 & 164.53 & 316.07 & 316.07 & 127.17 & 316.07 & 690.03 & 293.53 \\
& GIN & 152.13 & 168.35 & 182.95 & 201.64 & 169.71 & 156.16 & 251.23 & 200.45 & 251.65 \\
& MPNN & \textbf{8.33} & \textbf{7.51} & \textbf{32.08} & \textbf{4.56} & \textbf{51.90} & \textbf{27.18} & \textbf{51.04} & 124.89 & \textbf{41.73} \\
& HGT & 191.14 & 170.71 & 165.88 & 126.92 & 172.84 & 160.29 & 156.10 & 136.22 & 175.45 \\
& MHA & 197.36 & 30.27 & 96.56 & 27.10 & 59.58 & 52.42 & 138.48 & \textbf{80.22} & 60.72 \\
\midrule
\scriptsize{\multirow{2}{*}{\begin{sideways}Linear\end{sideways}}} & Node Set & 167.81 & 168.72 & 167.33 & 167.40 & 167.90 & 167.96 & 168.57 & 169.38 & 166.13 \\
& Edge Set & 181.44 & 181.21 & 181.18 & 181.32 & 180.86 & 179.44 & 181.08 & 181.68 & 181.40 \\
\bottomrule
\end{tabular}
}
\end{table*} 
\begin{table*}[htbp]
\centering
\caption{Degree Accuracy on all trees with 15 nodes.  We highlight the best performance per training task in columns.}\label{tbl:knn_degree_t15}
\footnotesize
\resizebox{\textwidth}{!}{\setlength{\tabcolsep}{4pt}
\begin{tabular}{ll|cccc|cc|ccc} 
\toprule
 & & \multicolumn{4}{c}{\textbf{Graph Tasks}} & \multicolumn{2}{c}{\textbf{Node Tasks}} & \multicolumn{3}{c}{\textbf{Edge Tasks}} \\
\cmidrule(lr){3-6} \cmidrule(lr){7-8} \cmidrule(lr){9-11}
& \textbf{Method} & \textbf{Node count} & \textbf{Edge count} & \textbf{Cycle check} & \textbf{Triangle counting} & \textbf{Node degree} & \textbf{Connected nodes} & \textbf{Reachability} & \textbf{Edge existence} & \textbf{Shortest path} \\
\midrule
\scriptsize{\multirow{5}{*}{\begin{sideways}Non-linear\end{sideways}}} & GCN & 53.57 & 55.15 & 55.24 & 25.91 & 25.91 & 54.86 & 25.91 & 11.08 & 36.51 \\
& GIN & 60.35 & 58.79 & 56.36 & 55.11 & 59.88 & 68.04 & 42.01 & 66.72 & 55.25 \\
& MPNN & \textbf{79.37} & \textbf{78.36} & 59.18 & \textbf{72.35} & 62.38 & 65.90 & 57.37 & 57.33 & 58.45 \\
& HGT & 54.88 & 55.33 & 55.34 & 58.65 & 54.33 & 58.84 & 57.27 & 57.43 & 55.34 \\
& MHA & 59.17 & 61.61 & 60.38 & 57.18 & 54.99 & 61.00 & 52.29 & 58.56 & 53.95 \\
\midrule
\scriptsize{\multirow{2}{*}{\begin{sideways}Linear\end{sideways}}} & Node Set & 65.64 & 66.32 & 65.93 & 66.10 & 66.13 & 65.95 & 66.28 & 66.22 & 65.82 \\
& Edge Set & 69.59 & {69.87} & \textbf{69.44} & 69.40 & \textbf{69.86} & \textbf{69.56} & \textbf{69.32} & \textbf{69.55} & \textbf{69.66} \\
\bottomrule
\end{tabular}
}
\end{table*} \begin{table*}[htbp]
\centering
\caption{Diameter Accuracy on all trees with 15 nodes. We highlight the best performance per training task in columns.}\label{tbl:knn_diameter_t15}
\footnotesize
\resizebox{\textwidth}{!}{\setlength{\tabcolsep}{4pt}
\begin{tabular}{ll|cccc|cc|ccc} 
\toprule
 & & \multicolumn{4}{c}{\textbf{Graph Tasks}} & \multicolumn{2}{c}{\textbf{Node Tasks}} & \multicolumn{3}{c}{\textbf{Edge Tasks}} \\
\cmidrule(lr){3-6} \cmidrule(lr){7-8} \cmidrule(lr){9-11}
& \textbf{Method} & \textbf{Node count} & \textbf{Edge count} & \textbf{Cycle check} & \textbf{Triangle counting} & \textbf{Node degree} & \textbf{Connected nodes} & \textbf{Reachability} & \textbf{Edge existence} & \textbf{Shortest path} \\
\midrule
\scriptsize{\multirow{5}{*}{\begin{sideways}Non-linear\end{sideways}}} & GCN & 50.77 & 50.36 & 49.54 & 25.97 & 25.97 & 50.01 & 25.97 & 6.77 & 26.64 \\
& GIN & 58.29 & 54.44 & 52.24 & 49.41 & 51.47 & 59.62 & 24.11 & 58.77 & 46.27 \\
& MPNN & 54.24 & 54.68 & 54.97 & 59.29 & \textbf{67.65} & \textbf{63.80} & 54.13 & 52.05 & 59.48 \\
& HGT & 57.15 & 54.88 & 54.90 & 57.58 & 57.05 & 65.22 & 54.51 & 58.70 & 53.07 \\
& MHA & 53.95 & 56.63 & 60.41 & 54.62 & 53.39 & 56.07 & 52.85 & 55.17 & 51.70 \\
\midrule
\scriptsize{\multirow{2}{*}{\begin{sideways}Linear\end{sideways}}} & Node Set & \textbf{61.89} & \textbf{62.68} & \textbf{62.74} & \textbf{62.36} & 61.99 & 61.93 & \textbf{62.34} & \textbf{62.49} & \textbf{62.40} \\
& Edge Set & 56.57 & 56.19 & 56.27 & 56.83 & 56.25 & 56.53 & 56.31 & 56.72 & 56.84 \\
\bottomrule
\end{tabular}
}
\end{table*} \begin{table*}[htbp]
\centering
\caption{Shortest path MSE on all trees with 15 nodes. We highlight the best performance per training task in columns.}\label{tbl:knn_sp_length_t15}
\footnotesize
\resizebox{\textwidth}{!}{\setlength{\tabcolsep}{4pt}
\begin{tabular}{ll|cccc|cc|ccc} 
\toprule
 & & \multicolumn{4}{c}{\textbf{Graph Tasks}} & \multicolumn{2}{c}{\textbf{Node Tasks}} & \multicolumn{3}{c}{\textbf{Edge Tasks}} \\
\cmidrule(lr){3-6} \cmidrule(lr){7-8} \cmidrule(lr){9-11}
& \textbf{Method} & \textbf{Node count} & \textbf{Edge count} & \textbf{Cycle check} & \textbf{Triangle counting} & \textbf{Node degree} & \textbf{Connected nodes} & \textbf{Reachability} & \textbf{Edge existence} & \textbf{Shortest path} \\
\midrule
\scriptsize{\multirow{5}{*}{\begin{sideways}Non-linear\end{sideways}}} & GCN & 12.95 & 12.31 & 12.62 & 26.17 & 26.17 & 12.22 & 26.17 & 49.78 & 21.71 \\
& GIN & 9.57 & 10.69 & 11.32 & 11.88 & 11.03 & 8.37 & 19.35 & 9.76 & 14.39 \\
& MPNN & \textbf{4.19} & \textbf{4.54} & 9.82 & \textbf{4.92} & \textbf{6.87} & \textbf{6.10} & 11.06 & 12.10 & 11.01 \\
& HGT & 10.57 & 10.96 & 11.65 & 9.09 & 12.56 & 8.17 & 10.76 & \textbf{9.26} & 10.98 \\
& MHA & 10.49 & 9.88 & \textbf{9.51} & 11.22 & 12.75 & 10.52 & 13.31 & 10.09 & 12.78 \\
\midrule
\scriptsize{\multirow{2}{*}{\begin{sideways}Linear\end{sideways}}} & Node Set & 10.20 & 10.05 & 10.13 & 10.11 & 10.17 & 10.21 & 10.07 & 10.18 & 10.03 \\
& Edge Set & 9.92 & 9.87 & 9.92 & 9.93 & 9.88 & 9.88 & \textbf{10.01} & 9.91 & \textbf{9.87} \\
\bottomrule
\end{tabular}
}
\end{table*} 

\end{document}